\documentclass[letterpaper, 10 pt, conference]{ieeeconf}  

\IEEEoverridecommandlockouts                              

\makeatletter
\let\NAT@parse\undefined
\makeatother

\usepackage[dvipsnames,table,xcdraw]{xcolor}
\usepackage{graphics}
\usepackage{amsmath}
\usepackage{orcidlink}
\usepackage{siunitx}
\usepackage{amssymb}
\usepackage{tikz}
\usepackage{hyperref}
\usepackage{afterpage}
\usepackage{geometry}
\hypersetup{
    pdffitwindow = true, 	colorlinks   = true,	
    breaklinks   = true,	urlcolor     = Blue,
    linkcolor    = black,	citecolor    = Blue,
    pdftitle     = {Magnet-Aware MPC+WBC Controller for Legged Locomotion},
    pdfauthor    = {Jorge Play\'an Garai}
}
\usepackage[all]{hypcap} 
\usepackage[capitalise,nameinlink]{cleveref}
\usepackage{pgfplots}
\usepgfplotslibrary{groupplots}
\usepackage{pgf}

\pgfplotsset{width=10cm, compat=1.9}

\begin{document}

\afterpage{\newgeometry{
  top=20.1mm,
  bottom=15.2mm,
  left=16.9mm,
  right=16.9mm
}}

\bstctlcite{IEEEexample:BSTcontrol}
\title{\LARGE \bf
    Magnet-Aware Control of Legged Robots
}

\author{
    Jorge Play\'an Garai$^{1,2}$\orcidlink{0000-0002-6130-5540},
    Sverre Bekkelund Djuve$^{2}$\orcidlink{0009-0002-1965-8112},
    Christopher McGreavy$^{2}$\orcidlink{0000-0001-8424-1941},
    Majid Khadiv$^{1}$\orcidlink{0000-0001-9889-6543}%
    \thanks{$^{1}$Technical University of Munich (TUM), Munich, Germany}
    \thanks{$^{2}$Conseil Européen pour la Recherche Nucléaire (CERN), BE/CEM/MRO, Geneva, Switzerland}
}

\maketitle
\thispagestyle{empty}
\pagestyle{empty}

\begin{abstract}

Autonomous robots can increase uptime and reduce human exposure in Big Science facilities, but strong magnetic fields needed for their operation corrupt sensors and induce pose-dependent mechanical wrenches that destabilize robots and challenge conventional reactive controllers. This paper presents a control framework for modeling, estimating, and dynamically compensating for spatially varying magnetic wrenches acting on legged robots to improve robustness in these fields. We introduce a custom physics plugin for the MuJoCo simulator to model magnetic forces on rigid-body elements, alongside an inverse field-estimation framework to infer the latent magnetic field directly from quadruped dynamic responses and any number of sensor readings. Furthermore, we develop a Magnet-Aware Model Predictive Control (MPC) and Whole-Body Control (WBC) architecture that predicts and counteracts magnetic perturbations during locomotion to increase the range of magnetic fields in which the robot can operate. The effectiveness of the framework is validated through both simulation and physical hardware experiments. We show our method increases the the maximum rejectable disturbances from magnetic field in the force space by a factor of 2.5 and and between 1.6-2 times in torque space compared to a non-compensated system. Within the proposed magnetic field, this constitutes an increase in the area in which the robot can operate by $ \boldsymbol{29,34\%} $ of which $ \boldsymbol{10,84\%} $ would have previously caused an immediate collapse to non-compensated controllers. 

\end{abstract}

\section{INTRODUCTION}
\label{se::introduction}

\begin{figure}[t]
    \centering
    \includegraphics[width=\columnwidth]{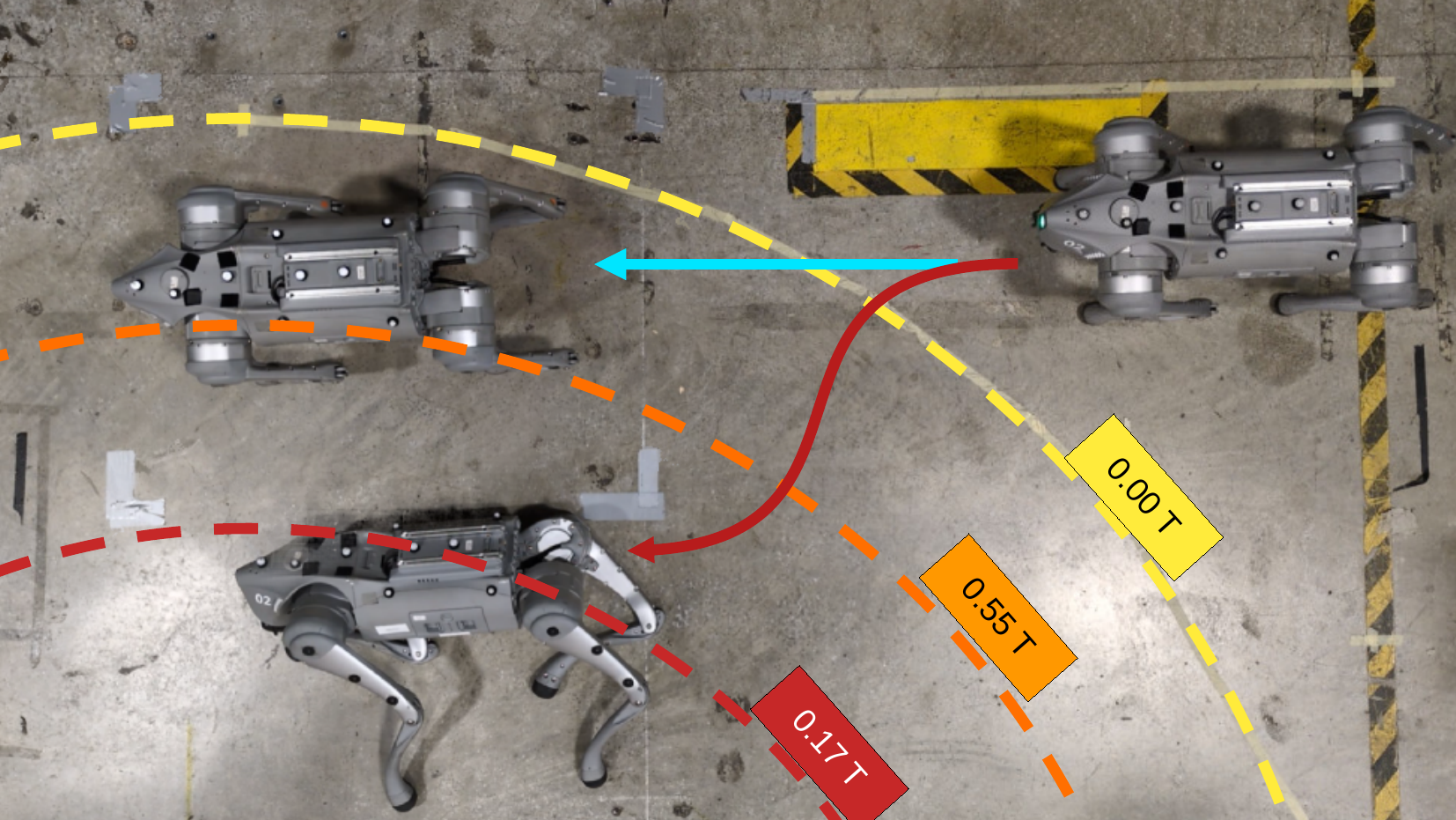}
    \caption{Robot moving through magnetic fields (dashed contour lines) experience induced disturbance wrenches due to their ferromagnetic components. Our method compensates for these disturbances (blue trace), allowing undisturbed movement and greater accessibility in magnetic field environments, where standard controllers would fail, leading to collapse (red trace)}
    \label{fig::diagram}
\end{figure}

Deploying autonomous robotic systems into Big Science facilities (such as high-energy physics and fusion energy experiments at CERN and ITER) is vital for automated inspection and maintenance. Operating within these sensitive environments requires high state estimation accuracy and robust disturbance rejection \cite{robotics_in_extreme_environments}. However, these facilities generate continuous, high-strength static magnetic fields reaching several Tesla ($\SI{}{\tesla}$) as a requirement for their operation, which jeopardises robotic operations. Robots unequipped to work in high magnetic fields are unable to withstand the magnetic fields in these facilities and can be pulled towards the magnetic field origin (Fig. \ref{fig::diagram}, red trace) causing the robot to fail and possibly damage the surrounding sensitive infrastructure. An adaptive, well-formulated compensation controller, however, could move through greater magnetic fields and reject higher induced disturbances using knowledge of the environment (Fig. \ref{fig::diagram}, blue trace), increasing the useful workspace of the robot.

All topologies are affected by these issues, from standard manipulators to more complex systems. However, Big Science facilities tend to be unstructured and multi-leveled, reducing the viability of wheeled mobile robots. Despite the added complexities of legged systems, these can overcome the presented challenges thanks to their enhanced mobility and adapting ground reaction forces to compensate existing perturbations.

Ferromagnetic components moving through a magnetic field experience significant forces and torques due to their interaction with the field. These induced perturbations impose continuous perturbations on the system that can threaten task completion. Additionally, electric brushless DC (BLDC) motors are most commonly equipped with Hall-effect encoders, which suffer from corrupted feedback and disrupted magnetic commutation \cite{brushless_dc_motors} \cite{hall_sensor_faults}, and will typically be the first point of failure in an electrically actuated system, forcing the motors to enter fault states and causing the collapse of the system. High-field environments also completely mask the Earth's magnetic field, rendering IMU magnetometers ineffective for yaw drift correction in state estimation \cite{magnetic_disturbance_imu} \cite{state_estimation_in_mag_fields}. Moving through steep magnetic flux gradients also induces parasitic currents in unshielded wiring (Faraday’s law \cite{Introduction_to_Electromagnetic_Compatibility}), threatening logic circuits and sensor signals.

Hardware modifications, such as magnetic-immune capacitive encoders \cite{capacitive_encoders}, and software adaptations such as magnetic-field SLAM \cite{magentic_field_navigation} \cite{magslam} mitigate some sensing issues, a holistic, environment-aware approach is needed for the control problem. In some approaches, robots are constructed without magnetic components to remove the control complexity completely \cite{6_dof_arm_for_mri}, but non-magnetically susceptible components drastically degrade the capabilities of a system and are limited in their applications. Momentum observers \cite{momentum_based_observer} \cite{momentum_observer_ground_force_estimation} \cite{momentum_based_kalman_filter} use the dynamic model of the robot and the momentum derivative to remove the need of noisy joint accelerations. A low pass filter is then added to create reactive strategies to compensate for, or measure, perturbations. However, this approach requires measurement or estimation of ground reaction forces (GRFs), and magnetic field induced disturbances are included in these GRF measurements and cannot be separated, leading to the disturbance force being masked within the GRF measurements. Projection approaches \cite{centroidal_state_estimation} \cite{khorshidi2025physically} project rigid-body dynamics of the system into the null space of the contact Jacobian to remove contact forces and isolate unmodeled forces. In magnetic environments, however, the non-linear relationship between magnetic fields and induced wrenches amplifies minor modeling inaccuracies into large disturbance readings. Furthermore, physical realities such as foot slip or contact compliance violate null-space assumptions and cause large residual spikes \cite{momentum_observer_issues}. Reactive or model-free disturbance rejection strategies with low-pass filtering aim for a high level of flexibility and adaptability, but, model-free approaches make decisions reactively based on sensor data within a moving window, leading to delays in compensating to magnetic field gradients, especially since such disturbances can not be detectable by these methods. Push recovery approaches to disturbance rejection \cite{push_recovery_quadruped} \cite{push_recovery_humanoid} assume external forces are transient, impulsive, or static, which is effective for short disturbances, but  magnetic wrenches form continuous vector fields that vary dynamically with spatial gradients and the robot's pose and joint configuration \cite{magnetic_materials}.

Existing disturbance estimation and control frameworks are therefore not suitable for these environments since they fail to guarantee the robust dynamic balance needed for autonomous operations. An effective control framework for legged robotic operation in a magnetic field needs to account for the complex dynamics of induced magnetic wrenches and should incorporate models of magnetic fields and their effects on susceptible ferromagnetic components of robot structures. 

To address these challenges, this paper focuses on modeling, estimating, and dynamically compensating for spatially varying magnetic wrenches on quadruped actuators. Critically, we will be assuming that the commutation in the motors is not done through Hall sensors but instead uses capacitive encoders as mentioned before. This will ensure our motors are reliable and robust even under magnetic fields. 
Our main contributions can be summarised as follows:

\begin{figure}
    \centering
    \includegraphics[width=\columnwidth]{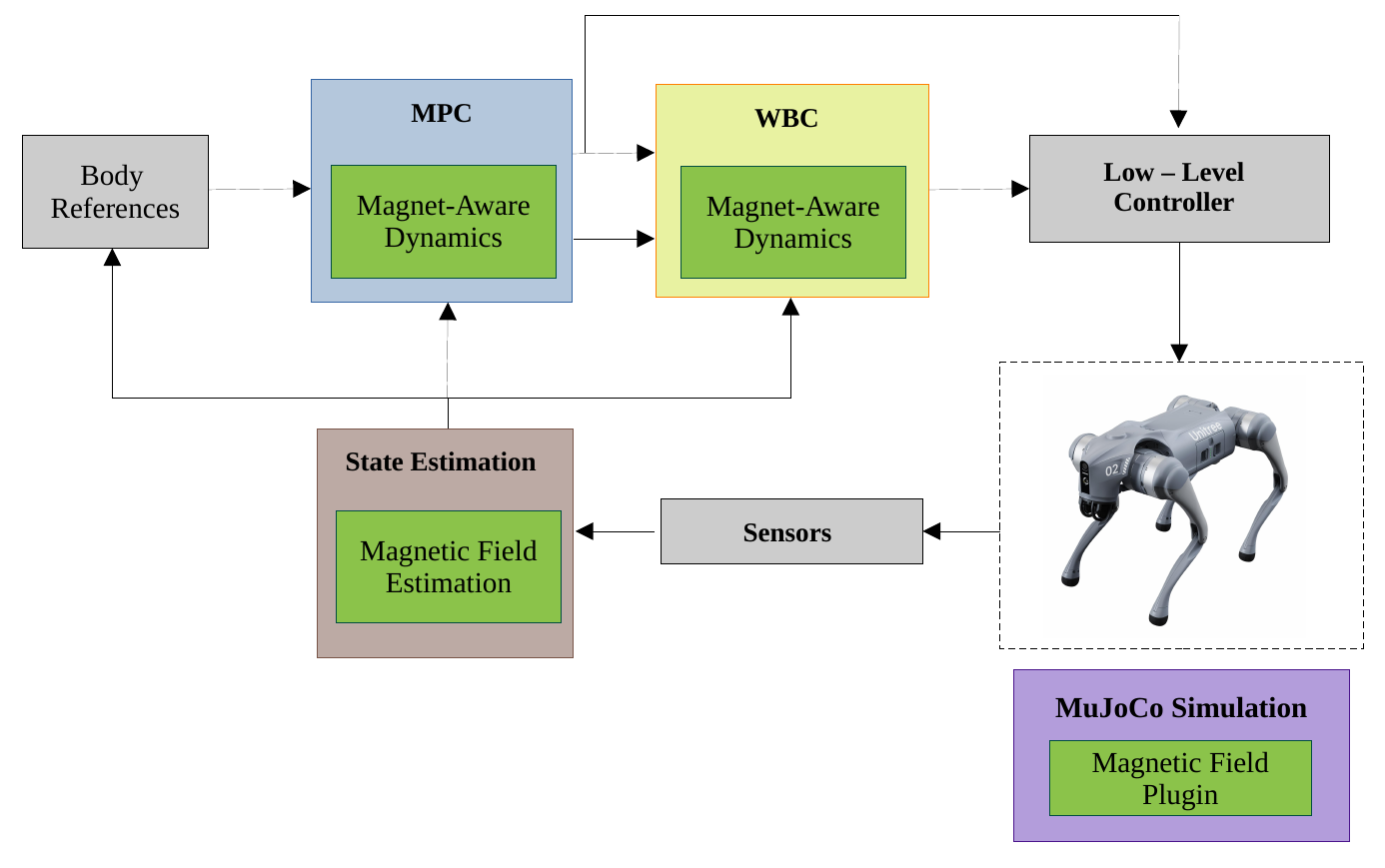}
    \caption{
        Block diagram of the proposed pipeline outlining the contributions. We present the pre-existing elements of the pipeline (grey), the components where we have applied modifications (blue, yellow, brown, purple), and the modifications themselves (green). 
    }
    \label{fig::block_diagram}
\end{figure}

\begin{enumerate}
    \item A custom MuJoCo \cite{mujoco} physics plugin that simulates induced magnetic forces and wrenches on rigid body elements based on the ferromagnetic components in the system.
    \item An inverse magnetic field estimation framework that infers local field distributions from legged system dynamics and sensor readings.
    \item A Magnet-Aware framework (Fig. \ref{fig::block_diagram}) that actively predicts and compensates for magnetic perturbations during dynamic locomotion.
\end{enumerate}

The remainder of this paper is organized as follows: Sec. \ref{se::background} details the background concepts and mathematical derivation of the magnetic wrench model. Sec. \ref{se::mujoco_implementation} describes the custom MuJoCo simulator implementation. Sec. \ref{se::dynamic_compensation} outlines the Magnet-Aware MPC+WBC dynamic compensation framework. Sec. \ref{se::magnetic_field_estimation} presents the magnetic field estimation algorithm. Finally, Sec. \ref{se::results} details simulation and physical hardware experiments evaluating controller performance.

\section{BACKGROUND}
\label{se::background}


\subsection{Notation}

We define the generalized configuration space of a floating-base legged system as $\boldsymbol{q}$, with its corresponding generalized velocity $\boldsymbol{\dot{q}}$ and acceleration $\boldsymbol{\ddot{q}}$ vectors defined as:

\begin{equation}
    \label{eq::cspace}
    \boldsymbol{q} =
        \begin{bmatrix}
            \boldsymbol{p} \\
            \boldsymbol{\phi} \\
            \boldsymbol{q_j}
        \end{bmatrix}
        \quad
        \boldsymbol{\dot{q}} =
        \begin{bmatrix}
            \boldsymbol{v} \\
            \boldsymbol{\omega} \\
            \boldsymbol{\dot{q}_j}
        \end{bmatrix}
        \quad
        \boldsymbol{\ddot{q}} =
        \begin{bmatrix}
            \boldsymbol{a} \\
            \boldsymbol{\dot{\omega}} \\
            \boldsymbol{\ddot{q}_j}
        \end{bmatrix}
\end{equation}

where $\boldsymbol{p} \in \mathbb{R}^3$ represents the position of the base frame with respect to the inertial frame $I$, and $\boldsymbol{v}, \boldsymbol{a} \in \mathbb{R}^3$ represent its linear velocity and acceleration. $\boldsymbol{\phi} \in \mathrm{SO}(3)$ denotes the orientation of the base frame, with $\boldsymbol{\omega}, \boldsymbol{\dot{\omega}} \in \mathbb{R}^3$ as its angular velocity and acceleration. The vectors $\boldsymbol{q}_{\mathrm{j}}, \boldsymbol{\dot{q}}_{\mathrm{j}}, \boldsymbol{\ddot{q}}_{\mathrm{j}} \in \mathbb{R}^{12}$ represent the joint positions, velocities, and accelerations, respectively.

We denote the magnetic flux density vector as $\boldsymbol{B} \in \mathbb{R}^3$ (measured in Teslas, $\mathrm{T}$) expressed in the inertial frame. Its spatial gradient tensor $\boldsymbol{G} = \nabla\boldsymbol{B}$ is represented as:

\begin{equation}
    \boldsymbol{B} =
        \begin{bmatrix}
            B_x \\
            B_y \\
            B_z
        \end{bmatrix}
    \quad
    \boldsymbol{G} =
        \begin{bmatrix}
            B_{xx} & B_{xy} & B_{xz} \\
            B_{yx} & B_{yy} & B_{yz} \\
            B_{zx} & B_{zy} & B_{zz}
        \end{bmatrix}
\end{equation}

In source-free regions, Maxwell's equations dictate that the field is curl-free ($\nabla \times \boldsymbol{B} = \boldsymbol{0}$) and divergence-free ($\nabla \cdot \boldsymbol{B} = 0$). Consequently, the gradient matrix $\boldsymbol{G}$ is symmetric ($B_{ij} = B_{ji}$) and traceless ($\mathrm{tr}(\boldsymbol{G}) = 0$).

\subsection{Dynamics of legged robots}

The floating-base rigid-body dynamics for a legged robot, such as a quadruped robot, can be defined as:

\begin{equation}
    \label{eq::dynamics}
    \boldsymbol{M}(\boldsymbol{q})\boldsymbol{\ddot{q}} + \boldsymbol{n}(\boldsymbol{q},\boldsymbol{\dot{q}}) = \boldsymbol{S}^\top\boldsymbol{\tau} + \boldsymbol{J}_c^\top\boldsymbol{\lambda}
\end{equation}

where $\boldsymbol{M} \in \mathbb{R}^{18 \times 18}$ is the generalized joint-space inertia matrix, $\boldsymbol{n} \in \mathbb{R}_{18}$ represents the non-linear effects of the system related to the Coriolis, centrifugal, and gravitational terms. $\boldsymbol{S} \in \mathbb{R}_{12 \times 18}$ is the selection matrix that maps our actuated joints torques $\boldsymbol{\tau} \in \mathbb{R}_{12}$ to the generalized coordinates. $\boldsymbol{J}_c \in \mathbb{R}_{3k \times 18}$ is the stacked contact Jacobian mapping contact forces $\boldsymbol{\lambda} \in \mathbb{R}_{3k}$ to the generalized coordinates, where $k$ is the number of feet in contact.

\subsection{Anisotropic Induced Magnetic Model}

To account for asymmetric and non-homogeneous magnetization arising from intrinsic material properties or geometric shape factors, an anisotropic magnetic model is adopted \cite{magnetic_materials}. Unlike the isotropic case where susceptibility is a scalar, anisotropic materials define susceptibility and permeability as tensors $\boldsymbol{\mu}, \boldsymbol{\chi} \in \mathbb{R}^{3 \times 3}$.

Geometric asymmetries are captured by the demagnetizing matrix $\boldsymbol{N} \in \mathbb{R}^{3 \times 3}$ expressed in the body frame $F$, leading to the apparent susceptibility tensor:

\begin{equation}
    \label{eq::apparent_susceptibility}
    \boldsymbol{\chi}_{\mathrm{app}} = \boldsymbol{\chi} (\boldsymbol{I}_3 + \boldsymbol{\chi} \boldsymbol{N})^{-1}
\end{equation}

Incorporating the element volume $V$ and the permeability of vacuum $\mu_0$, the magnetic polarizability tensor in the body frame $F$ is given by:

\begin{equation}
    {}_F\boldsymbol{\alpha} = \frac{V}{\mu_0} \boldsymbol{\chi}_{\mathrm{app}}
\end{equation}

This tensor is transformed into the inertial frame $I$ via the rotation matrix ${}_I\boldsymbol{R}_F$:

\begin{equation}
    {}_I\boldsymbol{\alpha} = {}_I\boldsymbol{R}_F {}_F\boldsymbol{\alpha} {}_I\boldsymbol{R}_F^\top
\end{equation}

Assuming a uniform field across the volume, the induced magnetic moment $\boldsymbol{m}$ expressed in the inertial frame (dropping frame prefixes for brevity) is defined as:

\begin{equation}
    \boldsymbol{m} = \boldsymbol{\alpha} \boldsymbol{B}
\end{equation}

Because $\boldsymbol{m}$ is generally non-collinear with the external magnetic field $\boldsymbol{B}$ in anisotropic elements, the cross product is non zero, resulting in both a magnetic gradient force $\boldsymbol{F}_m = \nabla(\boldsymbol{m} \cdot \boldsymbol{B})$ and a net magnetic torque $\boldsymbol{\tau}_m = \boldsymbol{m} \times \boldsymbol{B}$. Combining these contributions yields the complete magnetic wrench model $\boldsymbol{W}_m$:

\begin{equation}
    \label{eq::final_magnetic_wrench}
    \boldsymbol{W}_m =
    \begin{bmatrix}
        \boldsymbol{F}_m \\
        \boldsymbol{\tau}_m
    \end{bmatrix} = 
    \begin{bmatrix}
        \boldsymbol{G} \cdot (\boldsymbol{\alpha} \boldsymbol{B})  \\
        (\boldsymbol{\alpha} \boldsymbol{B}) \times{\boldsymbol{B}}
    \end{bmatrix}
\end{equation}

where $\boldsymbol{G}$ is the magnetic field gradient matrix. The resulting wrench is quadratic with respect to the magnetic field $\boldsymbol{B}$ for torque and bilinear with respect to $\boldsymbol{G}$ and $\boldsymbol{B}$ for force.


\section{PHYSICS SIMULATION}
\label{se::mujoco_implementation}

To compensate the effects of magnetic fields on our system, an accurate, representative and reliable physics simulator is required to model the effects of the magnetic field on the robotic system. Because magnetic field interactions are rarely integrated into rigid-body physics engines, existing robotics simulation frameworks lack natively supported tools to simulate these effects. Consequently, as part of the contributions of this work, we implemented a custom plugin within the MuJoCo physics environment \cite{mujoco} following an standard formulation described in Sec. \ref{se::background} to simulate magnetic forces acting on passive ferromagnetic materials.

Our module, allows any rigid body within the simulation model to be designated as a magnetic material. Users have to specify intrinsic material properties, such as relative magnetic permeability, demagnetizing factors, and the element's volume, to achieve realistic field responses. Additionally, the plugin requires an analytical definition of the external magnetic field as a function of spatial coordinates, which is evaluated at the center of mass of each magnetic body to compute local field vectors and spatial gradients.

As initial test environments, we defined two spatial magnetic field distributions:
\begin{enumerate}
    \item A linear field profile defined by a base field $\boldsymbol{B}_0$ at the origin with a constant spatial gradient tensor $\boldsymbol{G}$.
    \item A spherical field profile of radius $r$, where the field magnitude varies quadratically from zero at the outer boundary to a peak magnitude $\boldsymbol{B}_{\max}$ at the sphere's center.
\end{enumerate}

\begin{figure}
    \centering
    \includegraphics[width=1\linewidth]{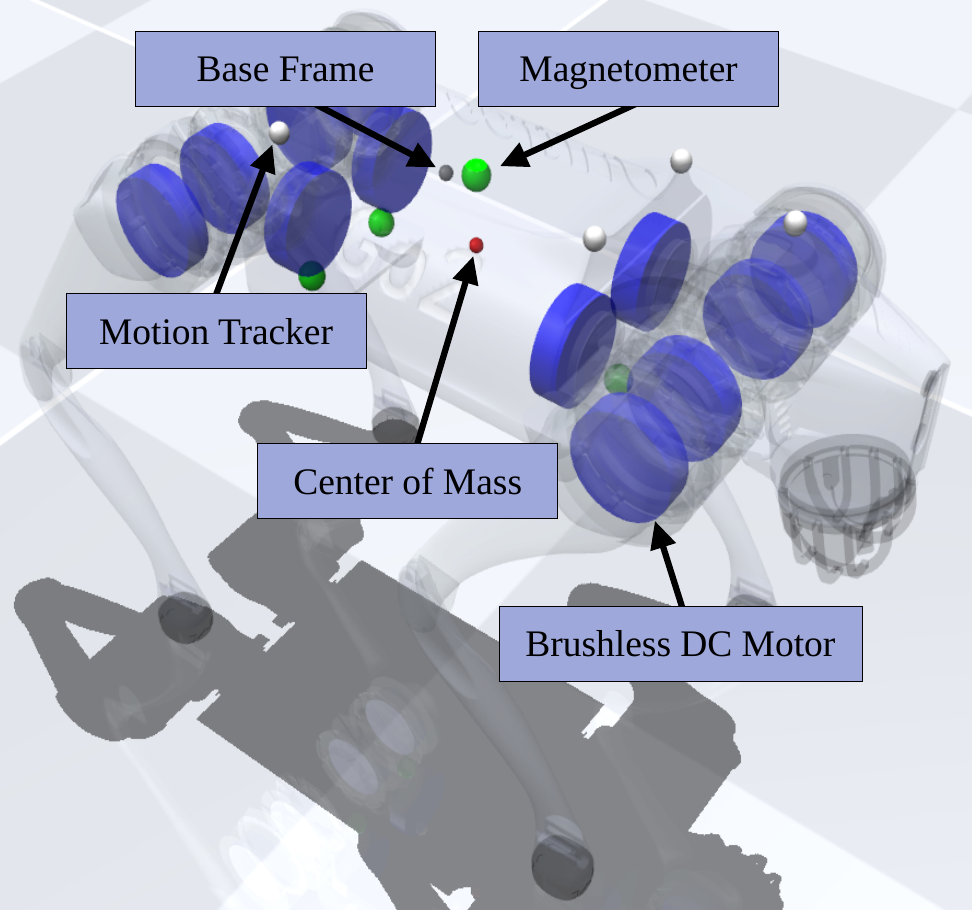}
    \caption{
        Simulation model of our quadruped robotic system. The 12 motor joints are modeled as induced magnetic cylinders (blue disks). Four magnetometers (green) are distributed across the chassis in a tetrahedral configuration. Motion tracker markers (white) correspond to the physical experimental platform. We also represent the base frame (gray) and center of mass (red).
    }
    \label{fig::simulation_go2}
\end{figure}

As shown in Fig. \ref{fig::simulation_go2}, all 12 joint motor assemblies of the Unitree Go2 platform are explicitly modeled as symmetrical cylinders matching the physical motor dimensions. As detailed in Sec. \ref{se::background}, the plugin requires intrinsic parameters for each body: relative permeability $\mu_r$, demagnetizing matrix $\boldsymbol{N}$, and body volume $V$, alongside environment field parameters ($\boldsymbol{B}_0$ and $\boldsymbol{G}$).
For our simulations, these parameters are configured as shown in Table \ref{tab:physical_values}.

\begin{table}[t]
    \centering
    \caption{Defined Magnetic Parameters values for Unitree GO-M8018-6 Motor}
    \label{tab:physical_values}
    \begin{tabular}{lccc}
        \hline
        \textbf{Field} & \textbf{Value} & \textbf{Notes} \\
        \hline
        N       & [0.2 0.2 0.6]                       & Normalized from \cite{demagnetizing_factors} \\
        V       &  $\SI{7.6969e-5}{\meter\cubed}$     & Obtained from motor CAD \\
        $\mu_r$ & 1000                                & Pure iron (worst-case scenario) \\
        \hline
    \end{tabular}
\end{table}

Because measuring the exact effective relative permeability of the composite motor assemblies is non-trivial, we assume a conservative worst-case scenario where each cylinder is composed entirely of pure iron ($\mu_r = 1000$). This maximizes the predicted magnetic wrench for a given field strength. Obtaining an accurate, representative model of the equivalent magnetic moment of motors remains an open issue for future research.


\section{DYNAMIC COMPENSATION}
\label{se::dynamic_compensation}

External magnetic fields and gradients induce parasitic wrenches across the quadruped's joint actuators. To maintain dynamic balance, these disturbance forces must be explicitly integrated into the robot's locomotion controller. As shown in Fig. \ref{fig::block_diagram}, we build upon a two-stage hierarchical control pipeline commonly used for quadrupedal locomotion \cite{unified_mpc_framework}: an MPC based on a kinodynamic model to optimize Ground Reaction Forces (GRFs), followed by a WBC formulated as a Quadratic Program (QP) to compute individual joint torques.

\subsection{Centroidal Dynamics in MPC}

The MPC computes the rate of change of centroidal momentum $\boldsymbol{h}$ as the sum of external forces and moments acting about the system's center of mass (CoM). We define the decision variable $\boldsymbol{x}_{mpc}(t) = [\boldsymbol{h}_{norm}^\top, \boldsymbol{q}^\top]^\top$ as the stacking of the normalized centroidal momentum and the c-space of our robot. We define the input control vector $\boldsymbol{u}_{mpc}(t) = [\boldsymbol{\lambda}^\top, \boldsymbol{\dot{q}}_j^\top]^\top$ as the stacking of the GRFs and the joint velocities. We augment the dynamics used in our optimal control problem by introducing the external magnetic wrenches computed in \eqref{eq::final_magnetic_wrench}:

\begin{equation}
    \label{eq::centroidal_dynamics_with_magnetic_wrench}
    \begin{bmatrix}
        \boldsymbol{\dot{p}}_c \\
        \boldsymbol{\dot{L}}_c \\
        \boldsymbol{\dot{q}}_b \\
        \boldsymbol{\dot{q}}_j
    \end{bmatrix}
    =
    \begin{bmatrix}
        \sum{\boldsymbol{\lambda}_i} + \sum{\boldsymbol{F}_{m,j}} + m\boldsymbol{g} \\
        \sum{(\boldsymbol{r}_i \times \boldsymbol{\lambda}_i}) + \sum{(\boldsymbol{\tau}_{m,j} + \boldsymbol{r}_{m,j} \times \boldsymbol{F}_{m,j})} \\
        \boldsymbol{A}_b^{-1}(\boldsymbol{q}) \left( \boldsymbol{h}_g - \boldsymbol{A}_j(\boldsymbol{q})\boldsymbol{\dot{q}}_j \right) \\
        \boldsymbol{\dot{q}}_j
    \end{bmatrix}
\end{equation}

where $\boldsymbol{\dot{p}}_c \in \mathbb{R}^3$ and $\boldsymbol{\dot{L}}_c \in \mathbb{R}^3$ represent the time derivatives of linear and angular centroidal momentum, respectively; $m$ is the total robot mass, and $\boldsymbol{g}$ is the gravity vector. The vectors $\boldsymbol{r}_i$ and $\boldsymbol{r}_{m,j}$ represent the positions of foot contact $i$ and joint motor $j$ relative to the CoM. The magnetic forces $\boldsymbol{F}_{m,j}$ and torques $\boldsymbol{\tau}_{m,j}$ are computed assuming known motor material properties and localized field values. $\boldsymbol{A}_b(\boldsymbol{q}) \in \mathbb{R}^{6 \times 6}$ and $\boldsymbol{A}_j(\boldsymbol{q}) \in \mathbb{R}^{6 \times n_j}$ are the base and joint submatrices of the centroidal momentum matrix $\boldsymbol{A}_g(\boldsymbol{q})$ evaluated at configuration $\boldsymbol{q}$. The resulting optimal control problem shown in \eqref{eq::mpc_optimal_control} is solved subject to standard operational constraints: friction cones, zero swing-foot forces, kinematic swing-foot tracking, and stance contact conditions.

\begin{equation}
    \label{eq::mpc_optimal_control}
    \begin{aligned}         
    \underset{\boldsymbol{x}_{mpc}(\cdot),\, \boldsymbol{u}(\cdot)}{\operatorname{min}} \quad 
        & \int_{t_0}^{t_f} \left\Vert{} \boldsymbol{r}(t) \right\Vert{}_{\boldsymbol{W}}^2 dt + \left\Vert{} \boldsymbol{r}_e(t_f) \right\Vert{}_{\boldsymbol{W}_e}^2 \\
    \operatorname{s.t.} \quad
        & \text{Centroidal Dynamics \eqref{eq::centroidal_dynamics_with_magnetic_wrench}} \\
        & \boldsymbol{x}_{mpc}(t_0) = \boldsymbol{x}_{mpc, 0} \\
        & \boldsymbol{\lambda}_{i}(t) \in \mathcal{F}_i(\mu) \quad \forall i \in \mathcal{C}(t) \\
        & \boldsymbol{\lambda}_{i}(t) = \boldsymbol{0} \quad \forall i \in \mathcal{S}(t) \\
        & \boldsymbol{v}_{\text{foot}, i}(t) = \boldsymbol{0} \quad \forall i \in \mathcal{C}(t) \\
        & \boldsymbol{n}^\top \boldsymbol{v}_{\text{foot}, i}(t) = v_{\text{des}, i}(t) \quad \forall i \in \mathcal{S}(t) \\
        & \boldsymbol{\underline{x}}_{mpc} \le \boldsymbol{x}_{mpc}(t) \le \boldsymbol{\overline{x}}_{mpc}, \\
        & \boldsymbol{\underline{u}}_{mpc} \le \boldsymbol{u}_{mpc}(t) \le \boldsymbol{\overline{u}}_{mpc}
    \end{aligned} 
\end{equation}

where $\boldsymbol{x}_{mpc}(t) = [\boldsymbol{h}^\top, \boldsymbol{q}^\top]^\top$ is the state vector stacking normalized centroidal momentum and generalized coordinates of the robot; $\boldsymbol{u}(t) = [\boldsymbol{\lambda}^\top, \boldsymbol{\dot{q}}_j^\top]^\top$ is the control input vector combining ground reaction forces $\boldsymbol{\lambda}$ and joint velocities; $\boldsymbol{r}(t) = \boldsymbol{r}(\boldsymbol{x}(t), \boldsymbol{u}(t), \boldsymbol{y}^{\text{ref}}(t))$ and $\boldsymbol{r}_e(t_f) = \boldsymbol{r}_e(\boldsymbol{x}(t_f), \boldsymbol{y}^{\text{ref}}(t_f))$ are the running and terminal tracking residuals penalizing state-input deviations against references $\boldsymbol{y}^{\text{ref}}$ under weight matrices $\boldsymbol{W}$ and $\boldsymbol{W}_e$; $\boldsymbol{x}_0$ is the initial state; $\mathcal{C}(t)$ and $\mathcal{S}(t)$ are the time-varying index sets for active stance and swing legs; $\mathcal{F}_i(\mu)$ defines the linear friction cone for leg $i$ under friction coefficient $\mu$; $\boldsymbol{v}_{\text{foot}, i}(t)$ is the cartesian linear foot velocity; $\boldsymbol{n}$ and $v_{\text{des}, i}(t)$ represent the ground surface normal vector and target normal velocity for swing leg $i$; and $\boldsymbol{\underline{x}}, \boldsymbol{\overline{x}}, \boldsymbol{\underline{u}}, \boldsymbol{\overline{u}}$ define state and input lower and upper bounds.

\subsection{Whole-Body Dynamics in WBC}

To map the trajectories from MPC to individual joint torques, we modify the rigid-body dynamics in \eqref{eq::dynamics}. Let $\boldsymbol{J}_{m,j} \in \mathbb{R}^{6 \times 18}$ denote the body Jacobian corresponding to the $j$-th motor body. Stacking these individual Jacobians into $\boldsymbol{J}_m \in \mathbb{R}^{6n \times 18}$ and the magnetic wrenches into $\boldsymbol{W}_m \in \mathbb{R}^{6n}$, the augmented floating-base dynamics are formulated as:

\begin{equation}
    \label{eq::dynamics_with_wrench}
    \boldsymbol{M}(\boldsymbol{q})\boldsymbol{\ddot{q}} + \boldsymbol{h}(\boldsymbol{q},\boldsymbol{\dot{q}}) = \boldsymbol{S}^\top\boldsymbol{\tau} + \boldsymbol{J}_{\mathrm{c}}^\top\boldsymbol{\lambda} + \boldsymbol{J}_m^\top\boldsymbol{W}_m
\end{equation}

In this formulation, the magnetic wrenches are projected into generalized coordinates via $\boldsymbol{J}_m^\top\boldsymbol{W}_m$. To evaluate these wrenches efficiently across predicted trajectories in MPC, spatial field distributions are approximated locally using a first-order Taylor series expansion.

Incorporating magnetic compensation directly into the rigid-body dynamics equations should significantly reduce pose tracking errors under magnetic field-induced disturbances. However, a key practical limitation is that compensation techniques cannot exceed the physical hardware limits. When forces and torques surpass friction limits or actuator capabilities the system will collapse regardless of the control strategy used. Consequently, in high-field environments, a combination of maximum joint torque limits, system weight, and perturbation direction determines the operational boundaries beyond which compensation alone can no longer prevent system collapse. The full optimization problem becomes

\begin{equation}
    \label{eq::wbc_optimal_control}
    \begin{aligned}
    \underset{\boldsymbol{\ddot{q}},\, \boldsymbol{\lambda},\, \boldsymbol{\tau}}{\operatorname{min}} \quad
        & \left\Vert{} \boldsymbol{A} \boldsymbol{x}_{wbc} - \boldsymbol{b} \right\Vert{}_{\boldsymbol{W}}^2 \\
    \operatorname{s.t.} \quad
        & \text{Rigid-body dynamics \eqref{eq::dynamics_with_wrench}} \\
        & \boldsymbol{J}_c(\boldsymbol{q})\boldsymbol{\ddot{q}} + \boldsymbol{\dot{J}}_c(\boldsymbol{q}, \boldsymbol{\dot{q}})\boldsymbol{\dot{q}} = \boldsymbol{0} \\
        & \boldsymbol{\lambda}_i = \boldsymbol{0} \quad \forall i \in \mathcal{S} \\
        & \boldsymbol{\lambda}_i \in \mathcal{F}_i, \quad \lambda_{z,i} \ge 0 \quad \forall i \in \mathcal{C} \\
        & \boldsymbol{\underline{\tau}} \le \boldsymbol{\tau} \le \boldsymbol{\overline{\tau}}
    \end{aligned}
\end{equation}

where $\boldsymbol{x}_{wbc}$ is the decision vector combining generalized accelerations, ground reaction forces, and joint torques; $\boldsymbol{A}$ and $\boldsymbol{b}$ are the task matrix and reference vector stacking base, swing leg, and force targets; $\boldsymbol{W}$ is the task-weighting diagonal matrix; $\boldsymbol{\dot{J}}_c(\boldsymbol{q}, \boldsymbol{\dot{q}})$ is the time derivative of the contact Jacobian matrix; and $\boldsymbol{\underline{\tau}}, \boldsymbol{\overline{\tau}}$ are the minimum and maximum motor torque limits.

\section{MAGNETIC FIELD ESTIMATION}
\label{se::magnetic_field_estimation}

As discussed in Sec. \ref{se::introduction}, while disturbance identification has been broadly studied using momentum observers, external magnetic fields introduce unique challenges. Quadruped robots are particularly challenging for momentum-based estimation due to their floating-base dynamics and complex contact interactions, which make external disturbance forces difficult to isolate from contact dynamics and gravity.

Consequently, we propose an optimization framework based on system identification techniques, extending the work of \cite{system_identification_todorov} to formulate a method to identify unknown magnetic fields using standard robot feedback. Using recorded joint torques $\boldsymbol{\tau}$ and kinematic trajectories from quadruped base motions, we evaluate candidate field parameters ($\boldsymbol{B}_0, \boldsymbol{G}$) to solve for the actual values present in the magnetic field. Forward simulation using our custom MuJoCo plugin is used for this. The original algorithm in \cite{system_identification_todorov} is used to identify the model parameters $\boldsymbol{P}$, by treating the magnetic field as part of the scene's model we can optimize only over the field or extend the problem to include the model parameters at the cost of increasing the complexity.

\begin{equation}
    \label{eq::mujoco_magnetic_estimation}
    [\boldsymbol{\ddot{q}}, \boldsymbol{s}] = \textbf{forward}(\boldsymbol{q}, \boldsymbol{\dot{q}}, \boldsymbol{\tau}, \boldsymbol{B}_0, \boldsymbol{G})
\end{equation}

where $\boldsymbol{s}$ denotes additional sensor observations.
Unlike \cite{system_identification_todorov}, which minimizes full joint acceleration errors $\boldsymbol{\ddot{q}}$, our approach uses a modified objective function. Obtaining reliable joint accelerations on physical hardware typically requires double-differentiating joint position encoders, which introduces severe high-frequency noise. Instead, we minimize the error between simulated and measured base accelerations $\boldsymbol{\ddot{q}}_{\mathrm{b}}$ and sensor measurements $\boldsymbol{s}$. An extra variation could be proposed to reintroduce joint accelerations in a weighted fashion if these are considered reliable enough.

\begin{equation}
    \label{eq:mujoco_magnetic_cost}
    \min_{\boldsymbol{B}_0,\boldsymbol{G}}
    \left(
    \left\|\boldsymbol{\ddot{q}}_{\text{b, meas}} - \boldsymbol{\ddot{q}}_{\text{b, sim}}\right\|_2^2
    +
    \left\|\boldsymbol{s}_{\text{meas}} - \boldsymbol{s}_{\text{sim}}\right\|_2^2
    \right)
\end{equation}

This formulation offers significant practical advantages: base accelerations are directly available from an onboard Inertial Measurement Unit (IMU), while angular accelerations can be reliably estimated via low-pass filtered finite differences of gyroscope signals.

\section{RESULTS}
\label{se::results}

We evaluate our proposed modeling, estimation, and compensation strategies through both simulation and physical hardware experiments. Since the magnetic field vector and its spatial gradient induce torques and forces simultaneously, assessing their isolated and coupled impacts on controller performance requires a structured experimental design.

The proposed MPC and WBC control pipelines were implemented using a custom in-house framework. Spatial rigid-body dynamics and kinematic Jacobians were computed using Pinocchio \cite{pinocchio_paper, pinocchio_github} and symbolic expressions generated via CasADi \cite{casadi}. The optimal control problem for the MPC stage was formulated and solved using Acados \cite{acados}, while the Whole-Body Control Quadratic Program (QP) was solved using qpOASES \cite{qpoases}.

\subsection{Estimation validation}

\begin{figure}[t]
    \centering
    \includegraphics[width=\columnwidth]{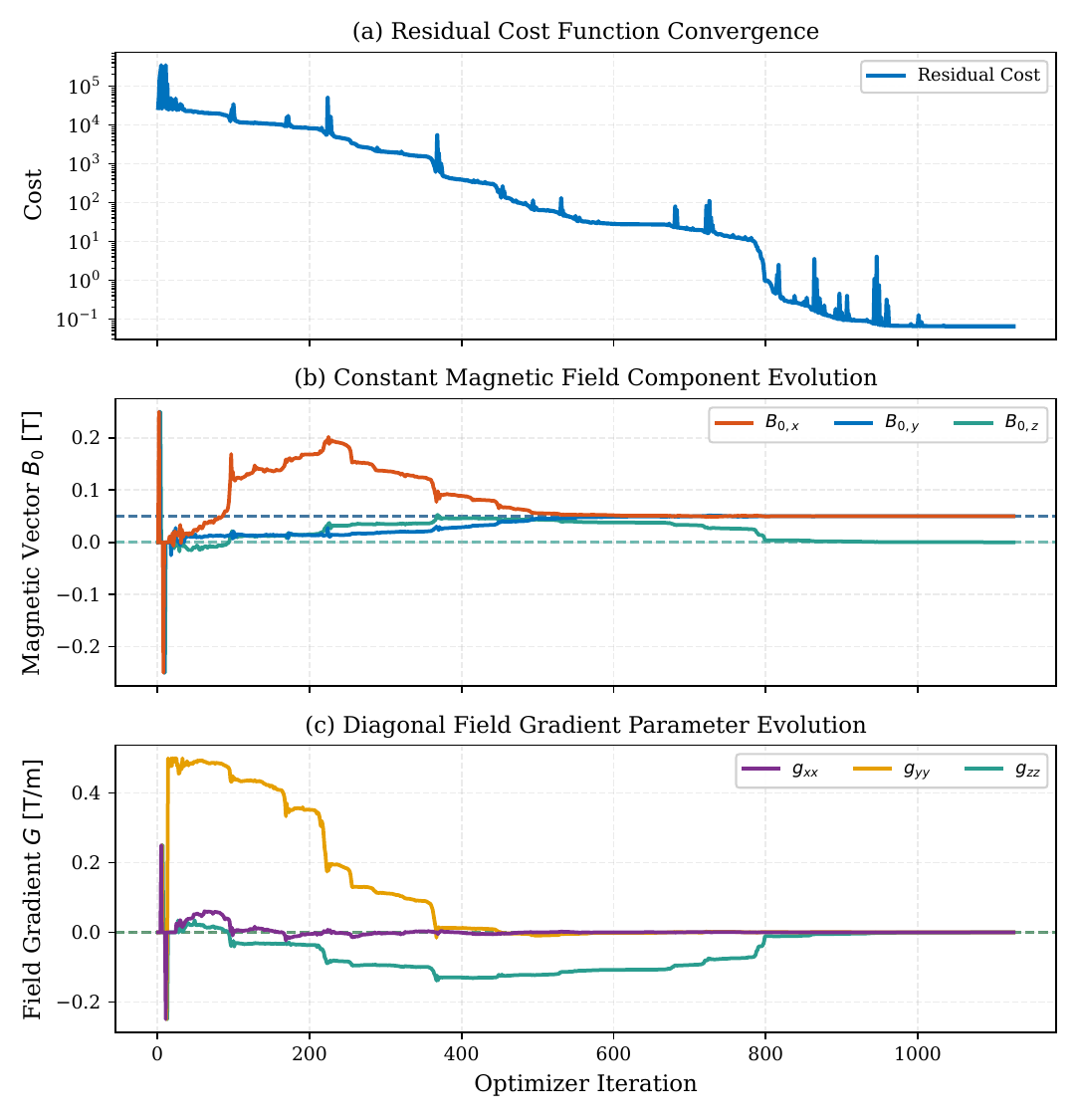}
    \caption{
        Convergence results for a static magnetic field. We record the robot as it moves through a scene with a magnetic field of norm $\SI{0.07}{\tesla}$. Optimizing over the results, we see that the algorithm eventually converges correctly to the proper magnetic field on the scene.
    }
    \label{fig::system_identificationc_convergence}
\end{figure}

We start validating the proposed magnetic field estimation algorithm. We run several simulations with different magnetic fields and with the robot walking the scene. As shown in Fig. \ref{fig::system_identificationc_convergence} the algorithm converges by using our plugin to iterate. However, high levels of noise and not enough sensor data can drastically affect the optimization. To improve the estimation, there should be as much data as possible. Another important point is the model of the field. In the presented case, we have used a linear field with a constant diagonal gradient and a field at the origin. However, for more complex fields, the models should scale accordingly or provide feasible approximations.

\subsection{Simulated Compensation Studies}

We first conduct simulation experiments in MuJoCo to establish the balance limits of both standard (uncompensated) and compensated control architectures under a constant magnetic field without spatial gradients. The magnetic field is aligned along the horizontal plane ($\boldsymbol{B} = [B_x, B_y, 0]^\top$), inducing a continuous yaw perturbation torque that attempts to rotate the quadruped's body by $\ang{45}$. 

Under standard control, the robot sustains a continuous magnetic disturbance torque up to $\SI{6}{\newton\metre}$. However, the uncompensated torque induces an unrecoverable yaw offset that increases with field strength, reaching a steady-state heading error of $\ang{10}$ at $B \approx \SI{0.1}{\tesla}$ before causing controller collapse at higher field strengths. The offset in Yaw starts being noticeable at $B \approx \SI{0.02}{\tesla}$.
In contrast, integrating active magnetic compensation allows the quadruped to maintain baseline heading and posture under field magnitudes that previously destabilized the robot. The compensated controller comfortably withstands continuous disturbance torques up to $\SI{10}{\newton\metre}$ (corresponding to field strengths of $B \approx \SI{0.18}{\tesla}$), reaching its saturation limit around $\SI{12}{\newton\metre}$.

To evaluate controller performance under coupled force and torque perturbations, we introduce a spatial magnetic gradient within a spherical volume of radius $r = \SI{2}{\meter}$. At the sphere's center, the field magnitude is set to $\boldsymbol{B}_{\mathrm{c}} = [0.3, 0.3, 0]^\top\,\mathrm{T}$. This field distribution generates a central alignment torque while inducing strong radially directed gradient forces $\boldsymbol{F}_m$. A quadratic gradient profile ensures that both forces and torques increase as the quadruped approaches the center.

\begin{figure}[t]
    \centering
    \includegraphics[width=\columnwidth]{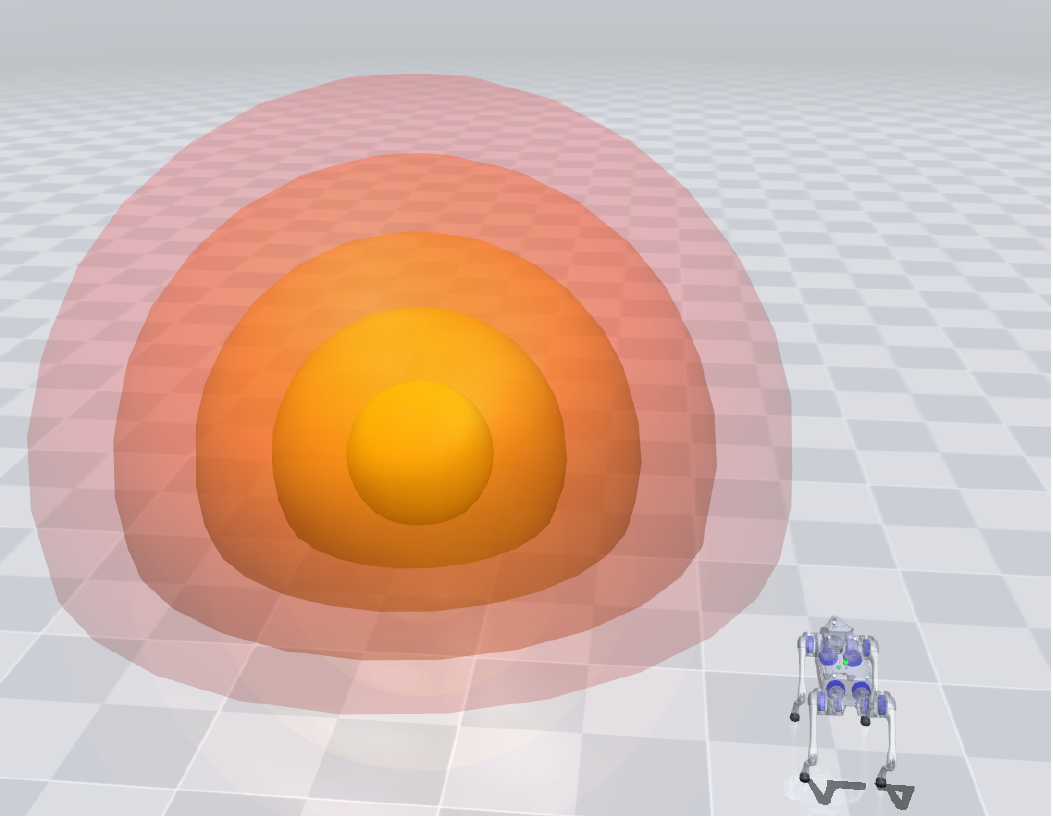}
    \caption{
        MuJoCo simulation scene for coupled field-gradient experiments. As the Unitree Go2 traverses the spherical magnetic region towards the epicenter, spatial field gradients induce increasing physical wrenches on the leg motor assemblies.
    }
    \label{fig::simulation_environment}
\end{figure}

 As shown in Fig. \ref{fig::simulation_environment}, where we present the simulation environment, the quadruped is set to track a linear trajectory passing through the magnetic sphere. By gradually moving the starting point along the Y axis we can increase or reduce the forces experienced by the quadruped allowing us to compare standard and compensated controllers.

\subsection{Hardware-in-the-Loop Validation}

Validating high-field magnetic compensation on physical quadrupeds poses significant practical challenges. Access to large-scale magnetic test facilities (such as those at CERN) is tightly restricted, and exposing physical hardware to unverified high-gradient fields risks structural or electrical damage.

\begin{figure}[t]
    \centering
    \includegraphics[width=\columnwidth]{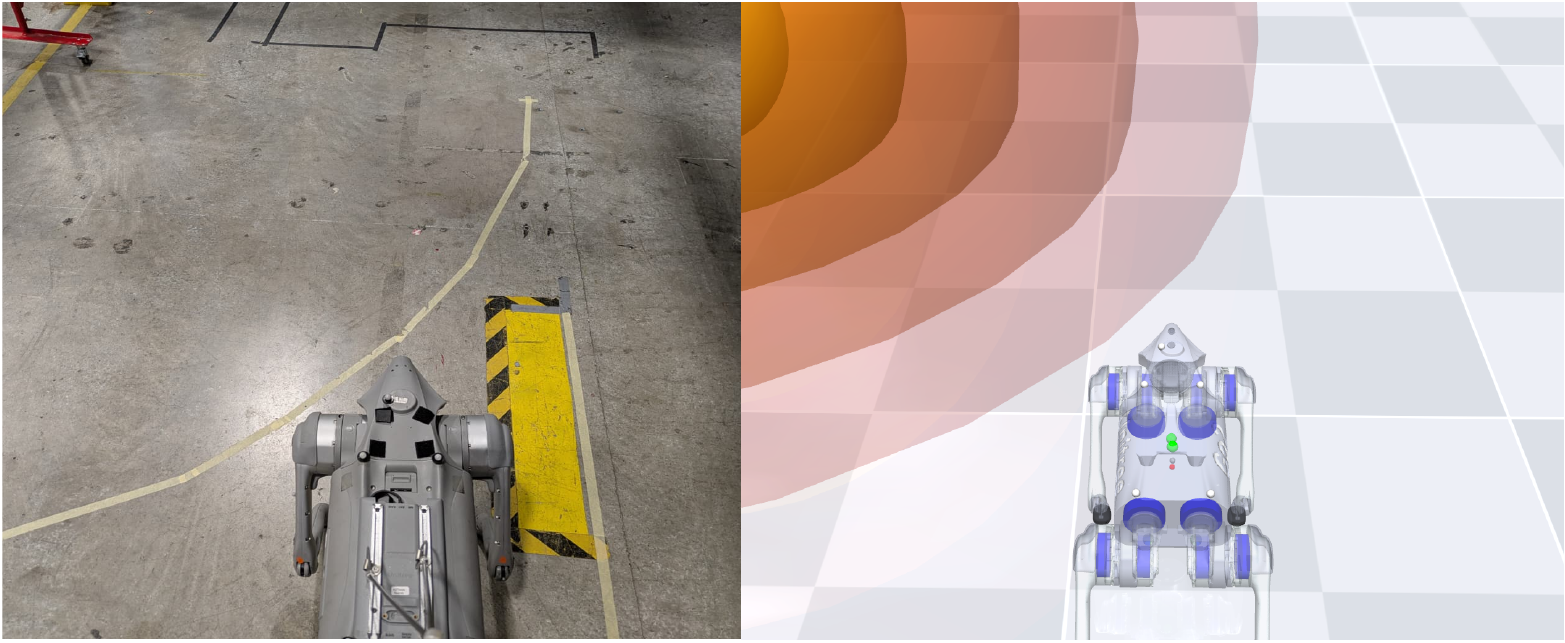}
    \caption{
        Experimental setup comparison. Right: Close-up of the MuJoCo simulation environment. Left: Physical laboratory testing ground with ground tape bounding the virtual magnetic field perimeter.
    }
    \label{fig::hardware_experimental_setup}
\end{figure}

To safely validate our framework on physical hardware, we adopt a Hardware-in-the-Loop (HIL) testing methodology. We execute the same simulated spherical magnetic field environment in MuJoCo concurrently with physical hardware tests on a Unitree Go2 quadruped \cite{unitree_go2}. At each control step, the physical quadruped's state $\boldsymbol{q}$ is updated in MuJoCo to evaluate simulated magnetometer readings. Four virtual magnetometers arranged in a tetrahedral configuration provide spatial field measurements, allowing local field gradients to be reconstructed. The comparative setup between simulation and hardware is illustrated in Fig. \ref{fig::hardware_experimental_setup}.

The simulated onboard magnetometers model the performance characteristics of commercial THM1176 three-axis Hall sensors \cite{metrolab_thm1176}, including realistic measurement noise and bias offset. To make the physical robot experience virtual magnetic perturbations during locomotion, the calculated magnetic wrench is injected directly into the hardware whole-body controller:
\begin{equation}
    \label{eq::hardware_field_simulation}
    \boldsymbol{M}(\boldsymbol{q})\boldsymbol{\ddot{q}} + \boldsymbol{h}(\boldsymbol{q},\boldsymbol{\dot{q}}) + \boldsymbol{J}_{m}^\top\boldsymbol{W}_{\mathrm{mujoco}}
    =
    \boldsymbol{S}^\top\boldsymbol{\tau} + \boldsymbol{J}_{\mathrm{c}}^\top\boldsymbol{\lambda} + \boldsymbol{J}_{m}^\top\boldsymbol{W}_{\mathrm{sensor}}
\end{equation}

where $\boldsymbol{W}_{\mathrm{mujoco}}$ represents the ideal spatial wrench generated by the MuJoCo environment, while $\boldsymbol{W}_{\mathrm{sensor}}$ represents the feedforward wrench computed from noisy magnetometer array feedback via a first-order Taylor series expansion.

Virtual wrench injection into the controller dynamics is necessary due to the floating-base nature of quadrupedal systems, where perturbation forces primarily act on underactuated base coordinates. Unlike fixed-base manipulators where disturbance torques can be directly commanded at individual joints, injecting magnetic wrenches through floating-base dynamics yields realistic joint torque distribution and contact force re-allocation.

\begin{figure*}[t]
    \centering
    \includegraphics[width=2\columnwidth]{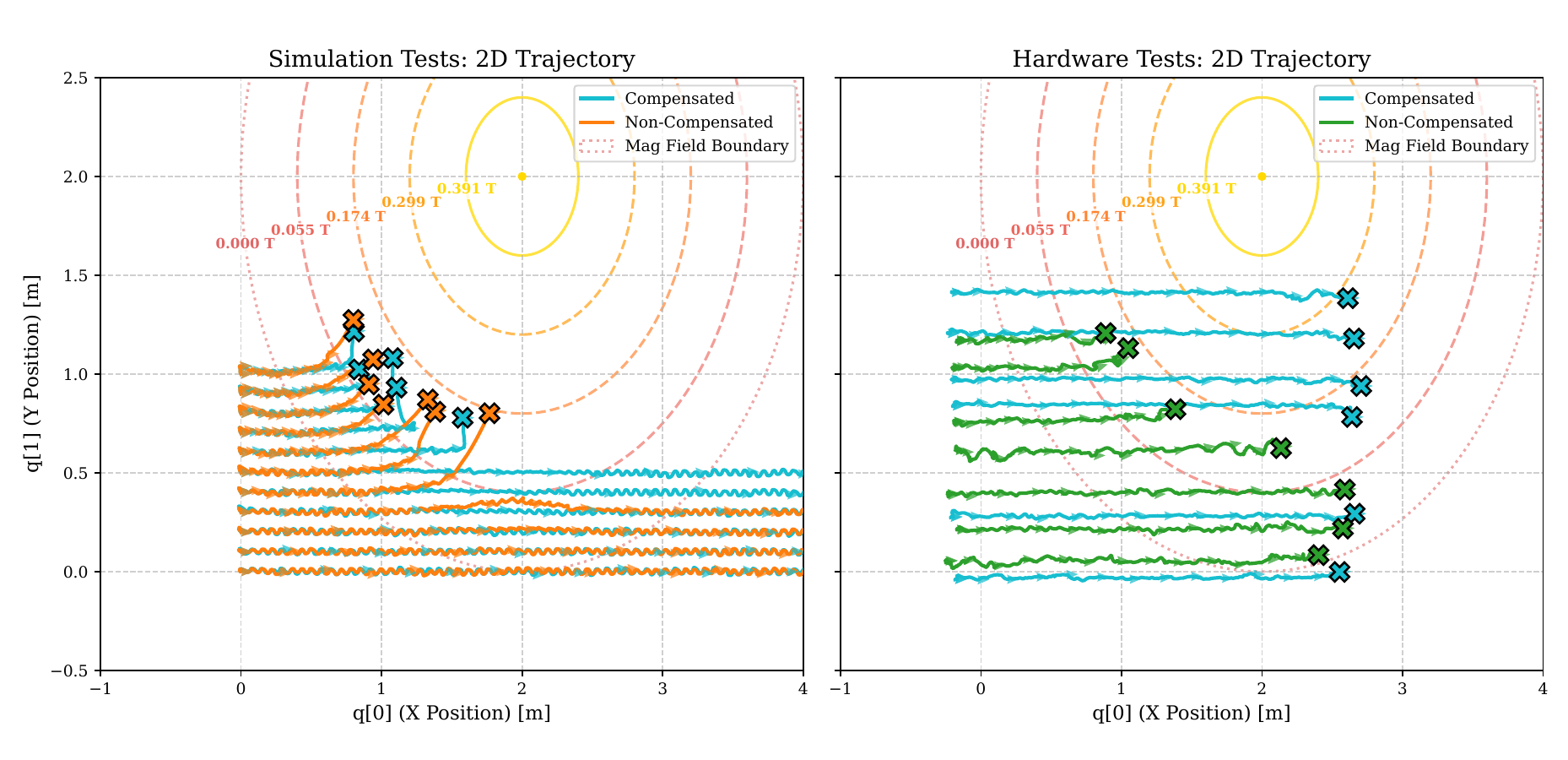}
    \caption{
        Experimental results comparing uncompensated and compensated controllers. Left: Simulation performance showing maximum trajectory penetration depths into the spherical magnetic field before collapse. Right: Hardware-in-the-loop experiments demonstrating improved dynamic balance and trajectory tracking under virtual field perturbations.
    }
    \label{fig::experiment_results}
\end{figure*}

\subsection{Performance Evaluation}

Fig. \ref{fig::experiment_results} shows that the uncompensated controller tolerates mild magnetic perturbations near the outer perimeter, but the robot loses balance as field strength increases, leading to foot slip and body collapse.
In full physics simulation, the standard controller exhibits initial loss of balance at field strengths between $\SI{0.02}{\tesla}$ and $\SI{0.03}{\tesla}$ (corresponding to net perturbation forces of $\approx \SI{15}{\newton}$), but manages to recover. However, when the local field reaches $\SI{0.05}{\tesla}$ (net force $\approx \SI{20}{\newton}$), locomotion failure occurs reliably. Plotting failure locations across multiple entry angles forms a distinct circular boundary around the field epicenter, representing the margin of recoverable disturbance.

Integrating active magnetic compensation significantly expands this operational area. The compensated quadruped withstands peak perturbation forces up to $\SI{50}{\newton}$, successfully traversing field magnitudes of $\SI{0.06}{\tesla}$ and $\SI{0.08}{\tesla}$ that previously caused collapse, and reaching peak transient fields of $\approx \SI{1.0}{\tesla}$ prior to actuator saturation. 
Pure magnetic field alignment torques induce yaw heading errors up to $\ang{5}$ under standard control, whereas compensated control suppresses heading deviation to $< \ang{1}$.

Finally, comparing physical HIL experiments with full physics simulation highlights an important structural distinction: in HIL tests, perturbation wrenches are injected numerically into the controller equations rather than applied as physical body forces acting on structural leg masses. Consequently, while in the full physics simulations the robot falls down when physical leg forces exceed friction limits, the HIL hardware setup successfully completes field traversals by generating feedforward joint torques without experiencing physical joint binding or external structural loading.

\begin{figure}[t]
    \centering
    \includegraphics[width=\columnwidth]{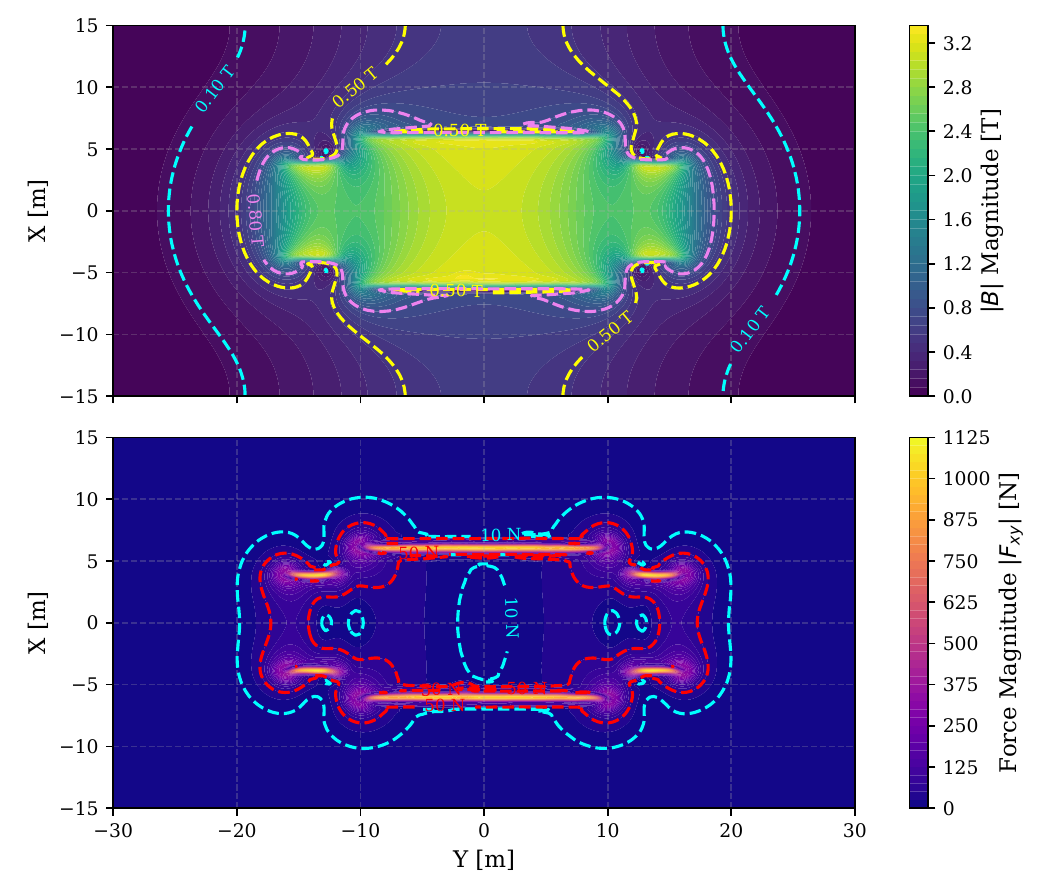}
    \caption{Simulations showing the resulting magnetic field from a proposed magnetic field for an FCC experiment. At top-down view is presented showcasing both the magnetic field norm (top) and the induced magnetic force norm expected (bottom). We present isodynamic lines at \SI{0.1}{\tesla} (cyan), \SI{0.5}{\tesla} (yellow), and \SI{0.8}{\tesla} (magenta). }
    \label{fig::magnet_operational_area}
\end{figure}

Fig. \ref{fig::magnet_operational_area} showcases the extended operational area achieved with this compensation within the premises of one of these detectors. Evaluating the margins observed in Fig. \ref{fig::experiment_results} marks a clear increase in the traversable workspace compared to the uncompensated baseline.

\section{CONCLUSIONS}

Operating legged robots in high-magnetic-field environments presents severe dynamic challenges due to localized parasitic forces and torques induced across ferromagnetic motors. This paper presented a comprehensive framework to model, compensate, and estimate these magnetic perturbations within a modern locomotion control architecture.

To evaluate these phenomena, we developed a custom magnetic simulation plugin for the MuJoCo physics engine, enabling accurate modeling of induced motor wrenches and spatial field distributions. By integrating these predicted magnetic wrenches directly into the floating-base dynamics of a two-stage MPC/WBC pipeline, the robot actively accounted for external field perturbations during locomotion. Both full physics simulations and Hardware-in-the-Loop (HIL) experiments on a Unitree Go2 quadruped demonstrated that the proposed compensation increases by a factor of 2.5 the disturbance force threshold tolerated by the platform prior to failures, and by a factor of 1.6-2 the disturbance torque. Consequently, this active compensation expanded the robot's safe operational range within the target high-field facilities by $ 29,34\% $, this percentage includes $ 10,84\% $ of the total area that would critically collapse the robot beforehand. 
Furthermore, we addressed the inverse field estimation problem. When onboard magnetometers are unavailable, noisy, or spatially constrained, the spatial magnetic field ($\boldsymbol{B}_0$) and gradient ($\boldsymbol{G}$) can be inferred through a system identification framework. By minimizing the discrepancy between simulated and measured base accelerations, the optimization pipeline identifies field parameters without requiring noisy joint acceleration signals.


\bibliographystyle{IEEEtran}
\bibliography{bibliography}

\end{document}